\documentclass[10pt,twocolumn,letterpaper]{article}

\usepackage{cvpr}
\usepackage{times}
\usepackage{epsfig}
\usepackage{graphicx}
\usepackage{amsmath}
\usepackage{amssymb}
\usepackage{multirow}
\usepackage{color}

\usepackage[breaklinks=true,bookmarks=false]{hyperref}

\cvprfinalcopy 

\def\cvprPaperID{****} 

\begin{document}

\title{HS-ResNet: Hierarchical-Split Block on Convolutional Neural Network}

\author{Pengcheng Yuan\footnotemark[1] , Shufei Lin\footnotemark[1] , Cheng Cui\footnotemark[1], Yuning Du, \\ Ruoyu Guo, Dongliang He, Errui  Ding, Shumin Han\footnotemark[2] \\
Baidu Inc.\\
\tt\small \{ yuanpengcheng01, linshufei, cuicheng01, duyuning, \\ 
\tt\small guoruoyu, hedongliang01, dingerrui, hanshumin \} @baidu.com
}


\maketitle

\renewcommand{\thefootnote}{\fnsymbol{footnote}} 
\footnotetext[1]{These authors contributed equally to this work.} 
\footnotetext[2]{Corresponding authors.}

\begin{abstract}
This paper addresses representational block named Hierarchical-Split Block,  which can be taken as a plug-and-play block to upgrade existing convolutional neural networks, improves model performance significantly in a network. Hierarchical-Split Block contains many hierarchical split and concatenate connections within one single residual block. We find multi-scale features is of great importance for numerous vision tasks. Moreover, Hierarchical-Split block is very flexible and efficient, which provides a large space of potential network architectures for different applications. In this work, we present a common backbone based on Hierarchical-Split block for tasks: image classification, object detection, instance segmentation and semantic image segmentation/parsing. Our approach shows significant improvements over all these core tasks in comparison with the baseline. As shown in Figure\ref{Figure1}, for image classification, our 50-layers network(HS-ResNet50) achieves 81.28\% top-1 accuracy with competitive latency on ImageNet-1k dataset. It also outperforms most state-of-the-art models. The source code and models will be available on: \url{https://github.com/PaddlePaddle/PaddleClas}
\end{abstract}

\section{Introduction}

In the past few years, Convolutional Neural Networks (CNNs) represent the workhorses of the most current computer vision applications, including image classification\cite{alexnet, urnet}, object detection\cite{fasterrcnn}, attention prediction\cite{attentiongrad}, target tracking\cite{targettracking}, action recognition\cite{action}, semantic segmentation\cite{deeplab,deeplabv3+}, salient object detection\cite{SOD}, and edge detection\cite{edgedetect}.

How to design a more efficient network architecture is the key to further improve the performance of CNNs. However, designing efficient architectures is becoming more and more complicated with the growing number of hyper-parameters (scale, width, cardinality etc.), especially when network is going deeply. In this paper, we rethink the dimension of bottleneck structure for network design. In particular, we consider the following three fundamental questions. (i) How to avoid producing abundant and even redundant information contained in the feature maps. (ii) How to promote the network to learn stronger feature presentations without any computational complexity. (iii) How to achieve better performance and maintain competitive inference speed.

\begin{figure}[t]
\centering
\includegraphics[width=0.5\textwidth]{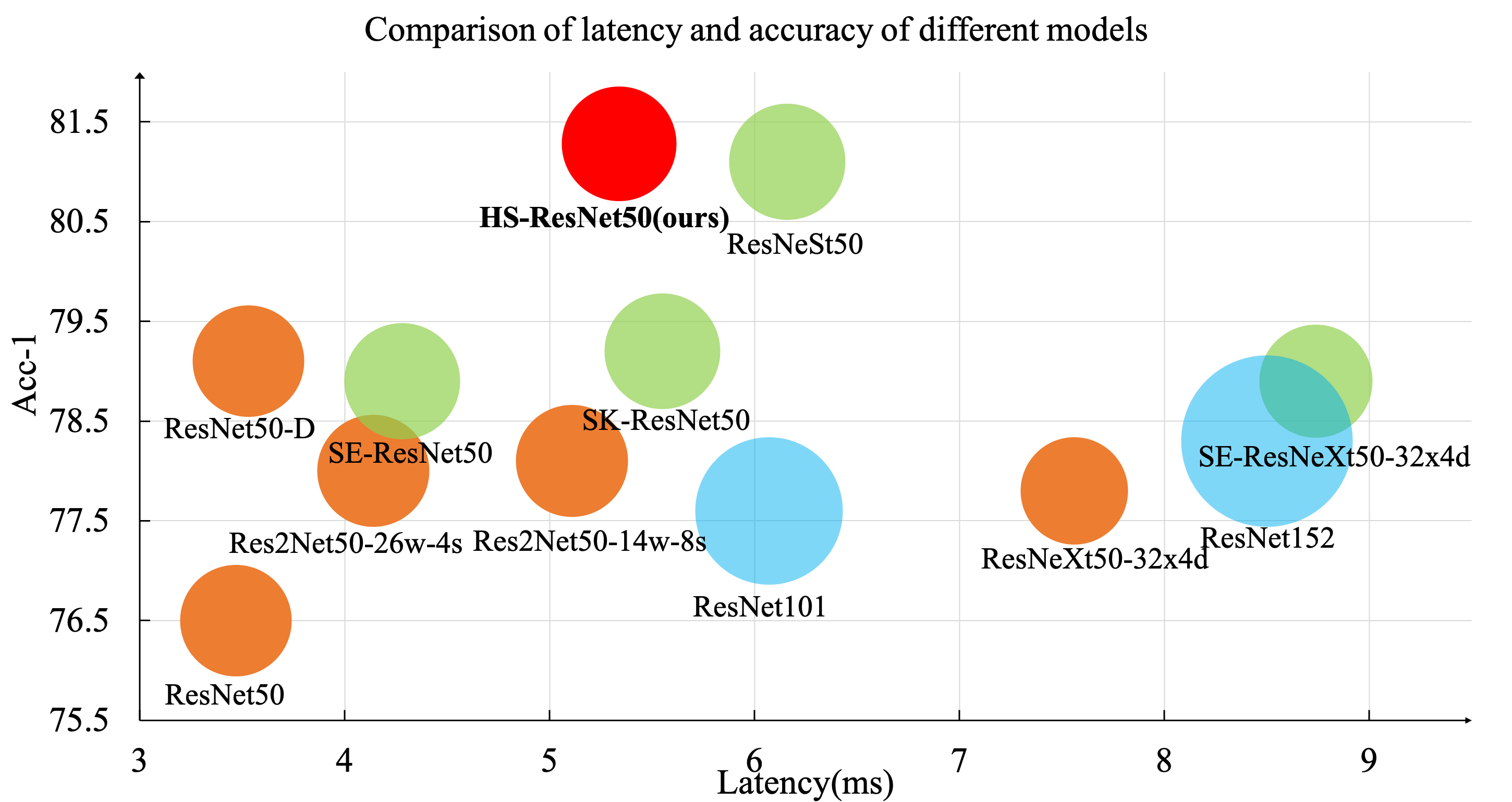} 
\caption{Comparing the accuracy-latency of different improved versions of ResNet50 models, the circle area reflects the params size of different models, latency test on T4/FP32 and batchsize=1. The orange circles represent different improvement strategies to improve the ResNet model, the green circles represent the use of different attention strategies, and the red circles are the HS-ResNet we designed. In order to make the indicators more intuitive, we added ResNet101 and ResNet152 represented by blue circles.}
\label{Figure1}
\end{figure}

\begin{figure*}[t]
\centering
\includegraphics[width=1.0\textwidth]{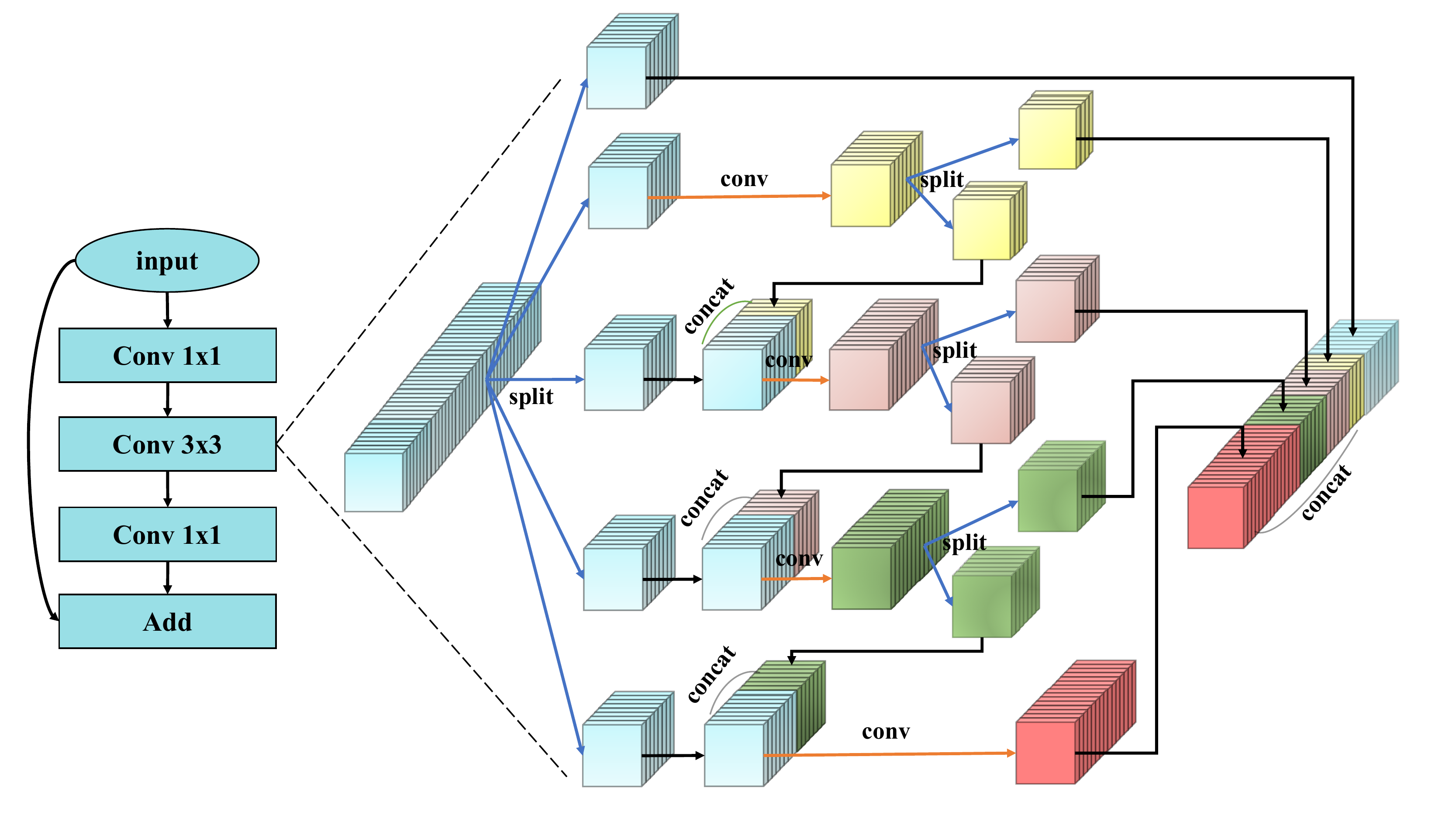} 
\caption{A detailed view of Hierarchical-Split block with $s$=5. \textbf{Split} means equally splitting in the channel dimension. \textbf{Conv} means $3\times3$ standard convolution + Batch normalization\cite{bn} + ReLU\cite{relu}. \textbf{Concat} means concatenating features in the channel dimension. This block only replaces the $3\times3$ convolution + Batch normalization + ReLU operation in the bottleneck of ResNet. The proposed Hierarchical-Split block can be taken as a plug-and-play component to upgrade existing convolutional neural networks.}
\label{Figure2}
\end{figure*}

In this paper, we introduce a novel Hierarchical-Split block to generate multi-scale feature representations. Specifically, an ordinary feature maps in deep neural networks will be split into $s$ groups, each with $w$ channels. As shown in Figure\ref{Figure2}, only the first group of filters can be straightly connected to next layer. The second group of feature maps are sent to a convolution of $3\times3$ filters to extract features firstly, then the output feature maps are splitted into two sub-groups in the channel dimension. One sub-group of feature maps straightly connected to next layer, while the other sub-group is concatenated with the next group of input feature maps in the channel dimension. The concatenated feature maps are operated by a set of $3\times3$ convolutional filters. This process repeats several times until the rest of input feature maps are processed. Finally, features maps from all input groups are concatenated and sent to another layer of $1\times1$ filters to rebuild the features. Meanwhile, we propose a network named HS-ResNet, which consists of several Hierarchical-Split blocks. Unlike Res2Net\cite{res2net},  we avoid producing the abundant and even redundant information contained in the feature maps and the network can learn richer feature representation. We summarize our main contributions as follows:

\begin{itemize}
\item We propose a novel Hierarchical-Split block, which contains multi-scale features. Hierarchical-Split block is very efﬁcient, and it can maintain a similar number of parameters and computational costs as the standard convolution. Hierarchical-Split block is very ﬂexible and extendable, opening the door for a large variety of network architectures for numerous tasks of computer vision.

\item We propose a network architecture for image classiﬁcation task that outperforms the baseline model by a signiﬁcant margin. Moreover, they are efﬁcient in terms of number of parameters and computational costs and evenly outperforms other more complex architectures.

\item We find that models utilizing a HS-ResNet backbone are able to achieve state-of-the-art performance on several tasks, namely: image classification\cite{alexnet}, object detection\cite{fasterrcnn}, instance segmentation\cite{maskrcnn} and semantic segmentation\cite{deeplab}. 
\end{itemize}

\section{Related Works}

To promote the capabilities of the model, current
works usually follow two types of methodologies. One is based on human design CNN architecture, another is based on Neural Architecture Search (NAS).
 
 \textbf{Human designed CNN architecture.}  The VGG\cite{VGG} exhibit a simple yet effective strategy of constructing very deep networks: stacking building blocks with the same dimension. GoogLeNet\cite{inceptionv1} constructs an Inception block, which includes four parallel operations: $1\times1$ convolution, $3\times3$ convolution, $5\times5$ convolution and max pooling. Then the output of these four operations are concatenated and fed to next layer. ResNeXt\cite{resnext} proposes new dimension named \textit{cardinality}, which operated by group convolutions, and considers that increasing cardinality is more effective than going deeper or wider when increase the capacity of network. SE-Net\cite{senet} introduces a channel-attention mechanism by adaptively recalibrating the channel feature responses. SK-Net\cite{sknet} brings the feature-map attention across two network branches. Res2Net\cite{res2net} improves the multi-scale representation ability at a more granular level. ResNeSt\cite{resnest} incorporates feature-map split attention within the individual network blocks.  PyConvResNet\cite{pyconv} uses pyramidal convolution, which includes four levels of different kernels sizes: $9\times9$, $7\times7$, $5\times5$, $3\times3$, to capture multiple scale features.

\textbf{Neural Architecture Search.} With the development of GPU hardware, main issue has changed from a manually designed architecture to an architecture that adaptively conducts systematic search for specific tasks. A majority of NAS-generated networks use the same or similar design space as MobileNetV2\cite{mobilenetv2}, including EﬃcientNet\cite{efficientnet}, MobileNetV3\cite{mbv3}, FBNet\cite{fbnet}, DNANet\cite{dnanet}, OFANet\cite{ofanet} and so on. The MixNet\cite{mixnet} proposed to hybridize depth-wise convolutions of diﬀerent kernel size in one layer. More earlier works such as DARTS\cite{darts} considered a hybrid of full, depth-wise and dilated convolution in one cell. However, NAS-generated networks relies on human-generated block like bottleNeck\cite{resnet}, Inverted-block\cite{mobilenetv2}. Our approach can also augment the search spaces for neural architecture search and potentially improve the overall performance, which can be studied in the future work.

\section{Method}

\subsection{Hierarchical-Split block}
The structure of the Hierarchical-Split block is shown in Figure\ref{Figure2}. After the $1\times1$ convolution, we split the feature maps into $s$ groups, denoted by $x_i$, and each group has equivalent $w$ regarding as the width of channels. Each $x_i$ will be fed into $3\times3$ convolutions, denoted by $\mathcal{F}_i\left( \right)$. The output feature maps of $\mathcal{F}_i\left( \right)$ are denoted by $y_i$. The most creative idea is to split the $y_i$ into two sub-groups, denoted $y_{i,1}$ and $y_{i,2}$. Then $y_{i,2}$ is concatenated with next group $x_{i+1}$, and then fed into $\mathcal{F}_{i+1}\left( \right)$, $\bigoplus$ means two feature maps are concatenated in the channel dimension. When finish operating all input feature groups, the channel dimension will be recovered by concatenating all $y_{i,1}$, which each $y_{i,1}$ has different channels. More channels $y_{i,1}$ contains, larger receptive field it gains. The output feature maps with smaller receptive field can focus on details, which are important for recognizing small objects or key parts of the objects, while concatenating more features from front groups would capture more larger objects.
In this work, we control $w$ and $s$ to limit the parameters or computational complexity of the HS-ResNet. Larger $s$ corresponds to stronger multi-scale ability, while larger $w$ corresponds to richer feature maps.

\begin{equation}
y_i=\left\{
\begin{array}{rcl}
x_i, & & {i = 1}\\
\mathcal{F}_{i}(x_{i} \bigoplus y_{i-1,2}), & & { 1< i <= s}\\
\end{array} \right.
\label{equ1}
\end{equation}

\subsection{Split and Concatenate operation}
The Hierarchical-Split block includes two stages split and concatenate operation. The first split operation turns the integrated feature maps $x$ into separate groups $x_i$, and each group has common width. The second split operation turns the alternate feature maps $y_i$ into two sub-groups, noted $y_{i,1}$ and $y_{i,2}$. It is worth noting that the width of $y_{i,1}$ and $y_{i,2}$ would be different when the width of $y_i$ is uneven. The design was inspired from GhostNet\cite{ghostnet}, the identity mapping for $y_{i,1}$ is to preserve the intrinsic feature maps, while the $y_{i,2}$ is used to capture more elaborate features. 

The first concatenation is used to widen the channels by concatenating the $y_{i-1,2}$ and $x_i$, enhancing the information flow between different groups. The second concatenation is merely combine all the $y_{i,1}$ and fed the output into the $1\times1$ convolution. We think concatenation is more effective than summation, which is used in Res2Net\cite{res2net}. Summation operation is likely to change, even destroy the feature representation, while concatenation will integrally maintain the feature representation.

\subsection{Analysis on Complexities}
We can approve that simply replacing the $3\times3$ convolution with our Hierarchical-Split block do not pay the price of increasing the number of parameters. Hierarchical-Split block has much less complexities than $k\times k$ convolution with the same $w$ and $s$. The computational complexity of $k\times k$ convolution is calculated by Equation \ref{conv:kxk}, while the computational complexity of each group is calculated by Equation \ref{conv:split-multi}. 
\begin{equation}
PARAM_{normal}=k \cdot k \cdot s \cdot w \cdot s \cdot w = k^{2}\cdot s^{2} \cdot w^{2} 
\label{conv:kxk}
\end{equation}
\begin{equation}
PARAM=\left\{
\begin{array}{rcl}
0, & & {i = 1}\\
k^{2} \cdot w^{2} \cdot (\frac{2^{s-1}-1}{2^{s-1}}+1), & & {1 < i <= s}\\
\end{array} \right.
\label{conv:split-multi}
\end{equation}

\begin{table*}[!htbp]
\centering
\begin{center}
\begin{tabular}{l|c|c|c|c} 
\hline
 Network & \#Params(M) & Top-1(\%) & Top-5(\%) &Inference Time(ms) \\
\hline
ResNet50\cite{resnet} & 25.56 & 76.50 & 93.00 & 3.47 \\
ResNet50-D \cite{bag_vd}& 25.58 & 79.12 & 94.44 & 3.53\\
SE-ResNet50-D\cite{senet} & 28.09 & 79.52 & 94.75 & 4.28 \\
Res2Net50-14w-8s\cite{res2net} & 25.72 & 79.46 & 94.70 & 5.40\\
ResNeSt50{*}\cite{resnest} & 27.50 & 81.02 & 95.42 & 6.69 \\
ResNeXt50-D-32x4d\cite{resnext} & 23.66 & 79.56 & 94.62 & 11.03  \\
SE-ResNeXt50-D-32x4d\cite{resnext} & 26.28 & 80.24 & 94.89 & 14.76 \\
\hline
\textbf{HS-ResNet50}(\textbf{Ours}) & 27.00 & \textbf{80.30} & \textbf{95.09} & 5.34 \\
\textbf{HS-ResNet50*}(\textbf{Ours}) & 27.00 & \textbf{81.28} & \textbf{95.53} & 5.34 \\
\hline
\end{tabular}
\end{center}

\caption{Validation accuracy comparison results of Hierarchical-Split block on ImageNet-1k\cite{imagenet} with other architectures, $*$ means adding extra data augmentations and training for 300 epochs. The times test on T4/FP32 and batchsize=1.}
\label{tabimageclassification}
\end{table*}

\begin{align}
PARAM &= k^{2} \cdot w^{2} \cdot \sum_{n=1}^{s-1}(\frac{2^{s-1}-1}{2^{s-1}}+1)\\
\notag
\Rightarrow    & = k^{2} \cdot w^{2}  \cdot  (\sum_{n=1}^{s-1}\frac{2^{s-1}-1}{2^{s-1}}+s-1)\\
     \notag
\Rightarrow      & < k^{2} \cdot w^{2} \cdot (s-1+s-1)\\
     \notag
\Rightarrow      & = k^{2} \cdot w^{2} \cdot (2 \cdot s-2)\\
     \notag
\Rightarrow      & < k^{2} \cdot w^{2} \cdot s^{2} \\ 
\notag
\Rightarrow      & < PARAM_{normal}
\end{align}


Apparently, our Hierarchical-Split block consume lesser resources than single convolution when the parameter of $s$ is going larger.

\section{Experiment}

\textbf{Training strategy.}We implement the proposed models using the PaddlePaddle framework. On the ImageNet-1k dataset\cite{imagenet}, each image are randomly cropped to $224\times224$ and randomly flipped horizontally. The GPU evaluation environment is based on T4 and TensorRT.

Batch Normalization\cite{bn} is used after each convolutional layer before ReLU\cite{relu} activation. We use label smoothing\cite{inceptionv3} and mixup\cite{mixup} as regularization strategies, and use SGD with weight decay 0.0001, momentum 0.9, and a mini-batch of 256. Our learning rates are adjusted according to a cosine schedule for training 200 epochs. 

In order to further improve the accuracy, we use Cutmix\cite{cutmix} instead of Mixup\cite{mixup}, and add data augmentations such as Rand Augmentation\cite{randaugment} and Random Erasing\cite{randomerasing}. Most importantly, we adjust the weight decay to 0.00004 for training 300 epochs.
 
\subsection{Image Classification.}
Table \ref{tabimageclassification} shows the top-1 and top-5 validation accuracy on the ImageNet-1K dataset\cite{imagenet}, which contains 1.28 million training images and 50k validation images from 1000 classes. For fair comparisons, all models in Table \ref{tabimageclassification} has the similar parameters. The HS-ResNet50 has an improvement of 1.2\% on top-1 accuracy over the ResNet50-D\cite{bag_vd} when these model are simply trained for 200 epochs. In addition, training HS-ResNet50 with more effective tricks, including  CutMix\cite{cutmix}, RandAugment\cite{randaugment}, Random Erasing\cite{randomerasing} and training for 300epochs, achieves 81.28\% on top-1 accuracy. Compared with ResNeSt50\cite{resnest}, the state-of-the-art CNN model, HS-ResNet50 not only outperforms by 0.26\% in terms of top-1 accuracy, but also is much efficient than ResNeSt50\cite{resnest}.

\begin{table}[!ht]
\begin{center}

\begin{tabular}{c|c|c|c}
\hline
 Method & Backbone & $\#LR$ & mAP(\%)\\
\hline
\multirow{5}{*}{\shortstack{Faster RCNN\cite{fasterrcnn} \\  + FPN\cite{fpn}}}& ResNet50\cite{resnet} & 1x  & 37.2  \\
& ResNet50-D\cite{bag_vd} & 1x  &38.9 \\
& ResNet101-D\cite{resnet} & 1x & 40.5    \\
& Res2Net50\cite{res2net} & 1x & 39.5    \\
& \textbf{HS-ResNet50} & 1x & \textbf{41.6}     \\
\hline
\end{tabular}
\end{center}

\caption{Object detection results on the COCO dataset\cite{mscoco}, measured using mAP@IoU=0.5:0.95 (\%).  $\#LR$ means learning rate scheduler.}
\label{tab:objectdetect}
\end{table}

\subsection{Object Detection.} 
For object detection task, we validate the HS-ResNet50 on the MS-COCO dataset\cite{mscoco}, which contains 80 classes. We use Faster R-CNN\cite{fasterrcnn} and FPN\cite{fpn} as the baseline method. Table \ref{tab:objectdetect} shows the object detection results. The entire
network is trained with stochastic gradient descent (SGD)
for 90K iterations with the initial learning rate being 0.02
and a minibatch of 16 images distributed on 8 GPUs. The
learning rate is divided by 10 at iteration 60K and 80K,
respectively. Weight decay is set as 0.0001, and momentum
is set as 0.9. Impressively, the HS-ResNet50 model improves the mAP on COCO\cite{mscoco} from 37.2\% to 41.6\%, and also outperforms ResNet101-D\cite{bag_vd} by 1.1\% at a faster inference speed.

\begin{table*}
\centering

\begin{center}
\begin{tabular}{c|c|c|c|c}
\hline
 Method & Backbone & $\#LR$ & Bbox mAP(\%) & Segm mAP(\%) \\
\hline
\multirow{5}{*}{\shortstack{Mask RCNN\cite{maskrcnn} \\ + FPN\cite{fpn}}}& ResNet50\cite{resnet}   & 2x & 38.7 & 34.7\\
& ResNet50-D\cite{bag_vd}  & 2x &  39.8  & 35.4 \\
& ResNet101-D\cite{resnet} & 2x & 41.4  & 36.8   \\
& Res2Net50\cite{res2net} & 2x & 40.9  & 36.2  \\
& \textbf{HS-ResNet50} & 2x & \textbf{43.1} & \textbf{38.0}     \\
\hline
\end{tabular}
\end{center}

\caption{Instance Segmentation results on the COCO\cite{mscoco} dataset, measured using bbox mAP@IoU=0.5:0.95 (\%) and segm mAP@IoU=0.5:0.95. $\#LR$ means learning rate scheduler. }
\label{tab:instanceseg}
\end{table*}

\begin{table*}
\begin{center}

\begin{tabular}{c|c|c|c|c|c}
\hline
 Network & Setting  & FLOPs(G) & Top-1(\%) & Top-5(\%)  & Inference Time(ms)\\
\hline
\multirow{4}{*}{\shortstack{HS-ResNet50\\(Same Params)}}& 18$w$-8$s$   & 11.6    &\textbf{81.43}    & 95.53    & 6.19\\
& 22$w$-7$s$   &12.3    &81.40  & 95.57    & 5.90  \\
& 28$w$-6$s$  &13.1    &81.28   & \textbf{95.61}    & 5.34   \\
& 40$w$-5$s$ &15.1    &81.11    & 95.40    & \textbf{5.28}   \\
\hline
\end{tabular}
\end{center}
\caption{Top-1 and Top-5 test accuracy (\%) of HS-ResNet-50 with different groups on the ImageNet-1k dataset. Parameter $w$ is the width of filters, and $s$ is the number of groups, as described in \ref{conv:kxk}. The evaluation environment is based on T4/FP32 and batchsize=1.}
\label{tab:scalesandwidth}
\end{table*}

\subsection{Instance Segmentation.}

For instance segmentation task, we also validate the HS-ResNet50 on the MS-COCO dataset\cite{mscoco}, using Mask R-CNN\cite{maskrcnn} and FPN\cite{fpn} as the baseline method. The hyper parameters are the same as object detection except training for 180K iterations, and the learning rate is divided by 10 at iteration 120K and 160K, respectively. Table \ref{tab:instanceseg} shows the instance segmentation results. The HS-ResNet50 based model outperforms ResNet50-D\cite{bag_vd} by 2.6\%, and ResNet101-D\cite{bag_vd} by 1.2\%.

\subsection{Semantic Segmentation.}
For semantic segmentation task, we also evaluate the multi-scale ability of HS-ResNet on the semantic segmentation task on Cityscapes dataset\cite{cityscapes}, which contains 5000 high-quality labeled images. We use the Deeplabv3+\cite{deeplabv3+} as our baseline method. We use the backbone of ResNet50-D\cite{bag_vd} vs. HS-ResNet50, and follow all other implementation details of \cite{zhao2017pyramid}. As shown in Table \ref{tab:semantic}, HS-ResNet50 outperforms ResNet50-D by 1.8\% on mean IoU.

\begin{table}[!h]
\centering
\begin{center}

\begin{tabular}{c|c|c}
\hline
 Method & Backbone & mIoU(\%) \\
\hline
\multirow{2}{*}{\shortstack{Deeplabv3+\cite{deeplabv3+}}}& ResNet50-D\cite{bag_vd}   & 78.0\\
& HS-ResNet50 & 79.8    \\
\hline
\end{tabular}
\end{center}

\caption{Performances of semantic segmentation on Cityscapes\cite{cityscapes} validation dataset.}
\label{tab:semantic}
\end{table}

\subsection{Ablation Study} 
\textbf{Width and Groups.} Table \ref{tab:scalesandwidth} shows the effects of different groups on inference speed. For fair comparisons, we adjust groups and width of filters to control the parameters of model. The larger the number of groups is, the smaller the width of filters become, accordingly. Larger groups will achieve high performance on top-1 accuracy, but slower inference speed. There are two main time-consuming factors. One is that serial process pattern among input feature maps plays a dominant role in our Hierarchical-Split block. The other is that split operations also increases latency.


\section{Conclusion and Future work}
This work proposed a novel Hierarchical-Split block that contains multi-scale feature representations. We built the HS-ResNet with Hierarchical-Split blocks, achieving state-of-the-art across image classification, object detection, instance segmentation and semantic segmentation. Our Hierarchical-Split block is very flexible and extendable, and thus should be broadly applicable across vision tasks. In the future, the source code and more experiments about other downstream tasks, such as OCR, video classification and scene classification, will be available on: \url{https://github.com/PaddlePaddle/PaddleClas}






{\small
\bibliographystyle{ieee_fullname}
\bibliography{egpaper_final}
}

\end{document}